\documentclass[preprint,12pt]{elsarticle}

\usepackage{amsmath,amssymb}
\usepackage[colorlinks=true,citecolor={blue!70!black},linkcolor={blue!70!black},urlcolor={blue!70!black},pdfborder={0 0 0}]{hyperref}
\usepackage{graphicx}
\usepackage{booktabs}
\usepackage{makecell}
\usepackage{adjustbox}
\usepackage{float}
\usepackage[section]{placeins}
\usepackage{xurl}
\usepackage[top=2.2cm,bottom=2.2cm,left=2.2cm,right=2.2cm]{geometry}
\journal{Computer Vision and Image Understanding}

\makeatletter
\def\ps@pprintTitle{%
  \let\@oddhead\@empty\let\@evenhead\@empty
  \def\@oddfoot{\hfil\thepage\hfil}\let\@evenfoot\@oddfoot}
\makeatother

\makeatletter
\let\origsubsection\subsection
\renewcommand{\subsection}{\FloatBarrier\origsubsection}
\makeatother

\begin{document}

\begin{frontmatter}

\title{Action Transfer Across Character Body Plans\\ in Latent-Action World Models:\\ An Empirical Study of Animated Characters}

\author{Amarnath R}
\address{Department of Information Technology, United Institute of Technology, Coimbatore, India\\
\textit{amarnathresearch@gmail.com}\vspace{-26pt}}

\begin{abstract}
Latent-action world models learn actions from video without labels, and recent methods aim to share these actions across bodies. Transfer between 2D characters with different body structures remains underexplored. In this paper, we study whether an action learned from one character can make a character with a different body perform the same action. A cat, a blob and a human can all jump, but each moves its body in its own way. The usual transfer gain mixes body motion with screen movement and can assign a perfect score to a model that ignores its action. To address this, we introduce a transfer score that compares the correct action with wrong ones from the same source. We test five training objectives on a new synthetic benchmark of 2D cartoon characters with six body plans, the standard Sprites benchmark, and 431 selected animal and human animations from DeformingThings4D. With a fixed camera, a baseline that only copies the source's screen movement scores 0.71, as high as most learned models. In contrast, with a camera that follows the body, the best model reaches only 0.46 for an unseen animal family and 0.20 for humans. With this camera, a linear probe identifies the action class of unseen animal families with 66\% to 86\% balanced accuracy (chance 17\%). The learned actions therefore retain information about which action is performed, but the correct action closes only part of the prediction error gap for a new character.
\end{abstract}

\begin{keyword}
action transfer \sep body plan \sep cross-body generalization \sep latent-action world models \sep action-specific transfer score \sep 2D character animation
\end{keyword}

\end{frontmatter}

\section{Introduction}

\begin{figure}[!htbp]
\centering
\includegraphics[width=0.9\textwidth]{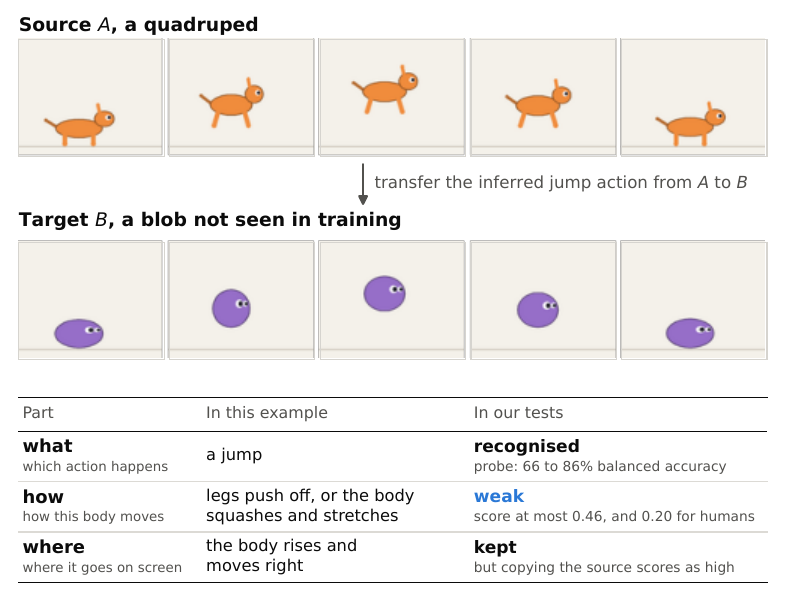}
\caption{A jump transferred from a quadruped to a blob not seen in training. The table lists the three parts of an action. The numbers in the last column summarise our tests on the animator-made animations and do not describe this example.}
\label{fig:problem}
\end{figure}

World models learn to predict future observations from the current state and an action. They are used to generate controllable videos and to learn representations that help agents plan \cite{ha2018,bruce2024}. However, many action-conditioned models rely on action labels during training, and most videos, such as films, cartoons and online clips, do not have them. Latent-action models address this limitation by learning actions directly from video. This is achieved through two connected models, an inverse model that infers the action from a pair of frames and a forward model that uses the inferred action to predict the later frame \cite{schmidt2024,adaworld}. These actions are inferred from visual changes between frames rather than specified through predefined action labels. Because no body-specific labels are needed, recent systems have explored whether such actions can also be transferred across different bodies \cite{lacwm,rynn,scar}. In animation, this could allow a motion learned from one character to be reused by another. For example, a model trained on videos of a jumping dog could make a cartoon rabbit, a snake or a blob jump, although it has never seen any of them jump before (Fig.~\ref{fig:problem}). The challenge is to transfer the action without simply copying the dog's movement, so that each character performs the jump in a way suited to its own body. A blob, for instance, has no legs and may express a jump by squashing and stretching.

Recent methods have tried to make latent actions transferable across different bodies. Some approaches limit the appearance or body information that the latent action can store \cite{claw,dila}, or align actions with the visual change they cause \cite{olaf}. Others show transfer between robot embodiments \cite{scar}, from a human hand to a robot gripper \cite{dila}, and from a moving person to a flying ball \cite{garrido2026}. However, these studies do not test whether the target body performs the action in its own way when its structure changes in a controlled way. Transfer between a quadruped, a snake and a blob is harder, because each body must perform the same action through different forms of movement. Garrido et al.~\cite{garrido2026} examined latent actions in everyday videos and found that, without a shared body, the learned actions could become local in the image and relative to the camera. Although they tested whether transferred actions could be recovered from the target video, they did not establish whether the target body could reproduce the intended motion. This also exposes a problem with the usual transfer score, which compares a transferred action against copying the previous frame. A model that largely ignores its action can therefore receive a high score. A model that only moves the character across the screen can also appear to transfer motion.

In this paper, we examine how well latent actions transfer between characters with different body plans. We consider three aspects of an action. These are what happens, how the body performs it, and where the character moves on the screen. To study these separately, we use two benchmarks for transfer across bodies and a third as a same-body control. The first is a synthetic set of 2D characters with six body plans, in which only the body structure changes and the other aspects of the motion stay the same. The second consists of animal and human animations from DeformingThings4D \cite{dt4d}, rendered as 2D side views to evaluate transfer on animator-made motion. The third is the standard Sprites benchmark \cite{sprites}, in which only the appearance of the characters changes. We compare five training objectives that share the same base network and training settings and differ in action size, noise and the losses that shape the action. We also train control models to separate the effects of action size and noise. Our study makes three contributions. First, we introduce a benchmark for evaluating transfer across different body plans under both fixed-camera and body-centred conditions. Second, we propose an action-specific transfer score that compares the correct transferred action with incorrect actions from the same source, and a baseline that copies only the source character's movement. Third, we show that action-class information can remain identifiable on unseen bodies even when body-specific motion transfers poorly, and examine how action size and training objectives affect this transfer.

\section{Related Work}

\subsection{Latent-action world models}

Latent-action models learn action representations from video without predefined action labels. LAPO \cite{schmidt2024} uses an inverse model to infer an action from two frames and a forward model to predict the later frame, and then uses the learned actions to pretrain policies. Genie \cite{bruce2024} uses a small set of discrete latent actions to make 2D platform games playable. AdaWorld \cite{adaworld} uses continuous latent actions to adapt a world model to new environments.

However, an action inferred from two frames may store more than the motion itself. It can also keep details about appearance, the layout of the scene or the source body. Several methods address this through their training objectives. CLAW \cite{claw} also runs the world model on the action alone, without the current frame, and uses gradient reversal on this path to discourage the action from carrying information about the future frame. Olaf-World \cite{olaf} averages the latent actions over a clip, projects the result into the feature space of a frozen video encoder and aligns it with the mean change in these features over the same clip, so that the same action means the same visual change in different clips. DiLA \cite{dila} lets the action drive only the structure of the predicted frame, and a separate pathway handles appearance. The methods differ in how they restrict or guide the information stored in the action. In this study, we adapt their main ideas to one common model and test how each affects transfer across body structures.

\subsection{Transfer across bodies}

Recent work has explored shared latent action spaces across several robots and agents. LAC-WM \cite{lacwm} trains a robot world model on a latent action space shared by many robots and adapts to new robots better than with explicit actions. RynnWorld-Latent \cite{rynn} drives one world model with latent actions taken from human, robot and simulation video. SCAR \cite{scar} reduces body and environment information in latent actions with an adversarial loss. It tests transfer between four robot embodiments in RoboTwin and between procedurally generated game environments in Procgen. Alvarez et al.~\cite{alvarez2026} supervise the similarity between latent actions with the similarity of the true actions and report better transfer between two bimanual robot arms in RoboTwin. They also report that a classifier identifies the robot from the latent actions with 100\% accuracy. In their split each task is shown by only one robot, so task and robot identity are confounded. DiLA \cite{dila} also moves a human hand motion onto a robot arm.

In the studies above, the bodies involved often share much of their structure, or the task provides a common action space. They do not show whether a transferred action can reproduce the body-specific motion of a character with a very different structure, in particular a character with no joints or skeleton.

Garrido et al.~\cite{garrido2026} study latent actions in everyday video, where the source and the target do not share a body. They show that the action of a walking person can make a ball move, and they check whether the transferred action can be inferred again from the resulting video. This shows that latent actions can cause changes in a different subject. However, recovering an action from the generated video does not show that the target has performed the intended body-specific motion. Our study addresses this by comparing the correct source action with wrong actions from the same source, and by including a baseline that transfers only the movement of the source across the screen.

\subsection{Motion retargeting, skeletons and 2D characters}

Motion retargeting transfers motion between characters by mapping it onto their skeletons. Skeleton-aware networks \cite{aberman2020} retarget between skeletons with different numbers of joints, and AnyTop \cite{anytop} generates motion for characters with any skeleton topology. Automatic rigging fits a skeleton to a 3D character \cite{pinocchio}, and more recent methods build skeletons for 3D characters with very different shapes \cite{dragon}.

For illustrated and 2D characters, pose estimators \cite{chen2022} and a system that animates children's drawings \cite{smith2023} assume a human-like skeleton. APES \cite{apes} takes a different approach and finds the moving parts of one game character from many of its sprites. These methods handle large changes in shape and skeleton topology. However, depending on the method, they need skeletons, rigs, pose labels or several examples of the target character. Such information is rarely available for 2D cartoon video, especially for characters whose bodies do not follow a skeleton at all.

Our setting differs from motion retargeting. We test whether a model can transfer an action learned from video pixels alone, without a skeleton, a rig or action labels for the target. This makes the question relevant to characters with very different bodies, including blobs and other characters without joints.

\subsection{Benchmarks for appearance and body variation}

Existing video benchmarks control appearance and motion in different ways. Sprites \cite{sprites} contains characters that share one body but differ in skin, clothing and hairstyle. It is therefore useful for testing transfer across changes of appearance, but it does not test transfer across body plans.

DeformingThings4D \cite{dt4d} provides animator-made 3D animations of many species, including humans. Its range of characters and motions makes it useful for studying transfer beyond a single body. In this work, we render these animations as flat 2D side views in one drawing style, so that the body changes while the appearance stays similar. We add a synthetic benchmark in which the body plan changes while the effort, direction and timing of each action are controlled. Together, these benchmarks test transfer under three conditions, a change of body plan with controlled motion, a change of appearance with a shared body, and animator-made motion across animal and human characters.

\section{Problem Formulation}

\subsection{Latent-action model}

We follow the standard latent-action set-up of LAPO \cite{schmidt2024} and AdaWorld \cite{adaworld}. A clip is a sequence of frames $x_1, \dots, x_T$. We sample every clip to $T = 16$ frames (12 on Sprites), which is enough to cover one complete action, such as a jump from take-off to landing. An inverse model $q$ infers an action between a frame and the frame $H$ steps later. A forward model $f$ then predicts the later frame from the earlier frame and the action,
\begin{equation}
z_t = q(x_t, x_{t+H}, x_{t+H} - x_t), \qquad \hat{x}_{t+H} = f(x_t, z_t), \qquad t = 1, \dots, T-H .
\end{equation}
Here $z_t$ is the action. It is a short list of numbers that acts as a code for the motion between the two frames, for example how much the body rises and how much the legs bend. The model is not given any predefined meaning for these numbers. It learns them from video alone, so $q$ must infer $z_t$ from the two frames. The length of this list sets how much the action can describe. One model in our study (M1) uses 128 numbers, and the others (M2 to M5) use 8 numbers (Section~\ref{sec:model}). We use $H = 3$, so that the two frames differ enough for the action to be visible. The actions of a clip $A$ are $z(A) = (z_1, \dots, z_{T-H})$. To transfer, we drive a target clip $B$ with the actions of a source clip $A$, step by step at the same frame index, and measure the prediction error
\begin{equation}
E(A \rightarrow B) = \sum_{t=1}^{T-H} \frac{\sum_{p} m_t(p)\, \lVert f(b_t, z_t(A))(p) - b_{t+H}(p) \rVert^2}{\sum_{p} m_t(p)} ,
\end{equation}
where $b_t$ is frame $t$ of $B$, $p$ runs over the pixels, and $m_t$ marks every pixel that belongs to the character in $b_t$, in $b_{t+H}$ or in both, so that the error covers where the character starts and where it ends. We count the error only on the character, because most of each frame is a still background that would make every prediction look good.

\begin{figure}[H]
\centering
\includegraphics[width=0.78\textwidth]{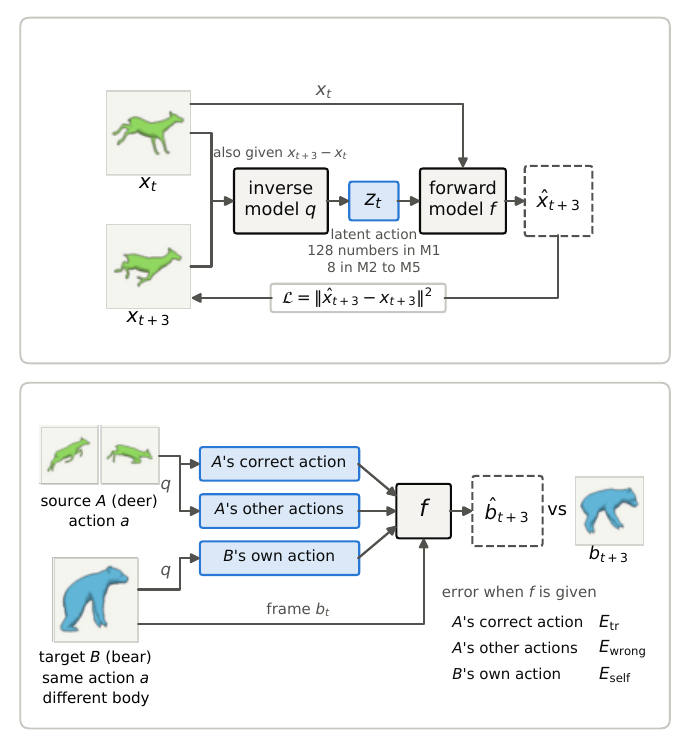}
\caption{The latent-action model. Top, training without action labels. Bottom, the transfer test. $f$ is run three times from the same target frame $b_t$, once with each kind of action, and each prediction is compared with the same real frame $b_{t+3}$. The three errors give the score $S$ of Section 3.4.}
\label{fig:model}
\end{figure}

Fig.~\ref{fig:model} shows the two stages. In training (top), the inverse model sees a deer at frames $t$ and $t+3$ and outputs an action $z_t$. The forward model receives frame $t$ and $z_t$ and predicts frame $t+3$. The loss compares this prediction with the real frame, so the forward model can only predict well if $z_t$ describes what changed between the two frames. In the transfer test (bottom), the trained models are fixed. We take actions from a source deer and give them to the forward model together with a frame of a target bear. We do this three times, with the bear's own action, the deer's correct action and the deer's other actions, and compare each prediction with the bear's real frame. These three errors are $E_{\mathrm{self}}$, $E_{\mathrm{tr}}$ and $E_{\mathrm{wrong}}$, from which we compute the score $S$ in Section 3.4.

\subsection{What, how and where}

\begin{figure}[H]
\centering
\includegraphics[width=\textwidth]{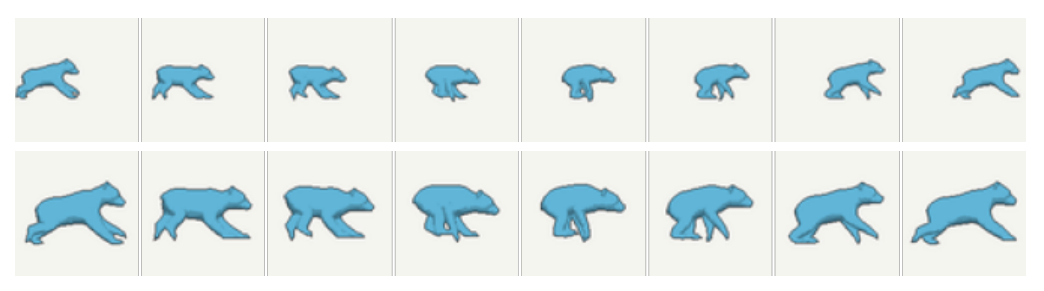}
\caption{One run of a bear with the fixed camera (top) and the body camera (bottom). The body camera removes the movement across the image, and the motion that remains is centred on the body.}
\label{fig:cameras}
\end{figure}

We measure the three parts of an action separately. To test where the character goes, we use a fixed camera and a baseline that moves the target character by the same amount as the source character and changes nothing else. To test how the body moves, we instead use a camera that follows the body. Such a camera removes the movement across the screen, so the motion that remains is centred on the body (Fig.~\ref{fig:cameras}). Rotation and changes of shape still remain, so this view does not isolate the movement of the limbs alone.

 To test what happens, we train a linear probe to read the action class from the actions of bodies that were not seen in training. The probe tests whether the action class can be identified from the learned actions. We also use a separate classifier to label the motion that a model generates, which checks the generated motion and not only the information in the action.

\subsection{Transfer pairs}

A transfer pair is a source character $A$ performing an action of class $c$ and a different target character $B$ performing an action of the same class. The real clip of $B$ is the correct answer. On the synthetic benchmark, $A$ and $B$ also share the effort, direction, start position and timing of the action, so the body is the only difference. The animator-made animations have no such control, so here the two clips only share the action class and cover the same part of the animation. Their timing can differ, and each source action drives the target frame with the same index, so a take-off may be paired with a landing. We test the effect of this with a separate best-shift check (Section~\ref{sec:eval}). We group pairs by whether the source and target bodies were seen in training. This gives seen to seen, seen to held-out, held-out to seen, and transfer to humans, who are never seen in training.

\subsection{Action-specific transfer score}\label{sec:score}

For each pair we compute four errors on the target clip $B_c$,
\begin{equation}
\begin{aligned}
E_{\mathrm{self}} &= E(B_c \rightarrow B_c), & E_{\mathrm{tr}} &= E(A_c \rightarrow B_c),\\
E_{\mathrm{wrong}} &= \frac{1}{|\mathcal{W}|} \sum_{c' \in \mathcal{W}} E(A_{c'} \rightarrow B_c), & E_{\mathrm{copy}} &= \sum_{t=1}^{T-H} \frac{\sum_{p} m_t(p)\, \lVert b_t(p) - b_{t+H}(p) \rVert^2}{\sum_{p} m_t(p)} ,
\end{aligned}
\end{equation}
where $\mathcal{W}$ is the set of other actions of $A$. This set is fixed for each source and is the same for every method. $E_{\mathrm{self}}$ uses $B$'s own action, $E_{\mathrm{tr}}$ uses $A$'s correct action, $E_{\mathrm{wrong}}$ uses $A$'s other actions, and $E_{\mathrm{copy}}$ predicts each frame by the frame $H$ steps earlier. All four use the same mask and the same averaging. The usual transfer gain compares the transferred action with copying,
\begin{equation}
G = \frac{E_{\mathrm{copy}} - E_{\mathrm{tr}}}{E_{\mathrm{copy}} - E_{\mathrm{self}}} .
\end{equation}
If a model ignores its action, then $E_{\mathrm{tr}} = E_{\mathrm{self}}$ and $G = 1$, a perfect score. We compare the correct action with wrong actions instead,
\begin{equation}
S = \frac{E_{\mathrm{wrong}} - E_{\mathrm{tr}}}{E_{\mathrm{wrong}} - E_{\mathrm{self}}} .
\end{equation}
$S = 1$ means the transferred action predicts $B$ as well as $B$'s own action. $S = 0$ means it is no better than the wrong actions on average, and $S < 0$ means it is worse. $S$ can also exceed 1 when the transferred action predicts $B$ better than its own action, and we do not clip it. $S$ is a relative score computed from prediction errors. A value of 0.46 means that the transferred action closes 46\% of the error gap between wrong actions and the target's own action. It does not mean that 46\% of the motion was copied. $S$ measures how much better the correct action predicts the target than wrong actions do. It is a measure of action-specific prediction, not a direct measure of how well limbs, contact or deformation are reproduced.

A simple example shows the difference between the two scores. For a model with $E_{\mathrm{self}} = 0.010$, $E_{\mathrm{tr}} = 0.014$, $E_{\mathrm{wrong}} = 0.020$ and $E_{\mathrm{copy}} = 0.030$, the correct action is better than the wrong ones but worse than $B$'s own action, which gives $S = 0.6$ and $G = 0.8$. A model that ignores its action gives the same error for every action, for example $E_{\mathrm{self}} = E_{\mathrm{tr}} = E_{\mathrm{wrong}} = 0.012$. It receives $G = 1.0$, a higher score than the first model, although it transfers nothing. Its $S$ is undefined, because the actions give no difference to measure. In a separate check, models trained for only one epoch get $G = 1.00$ and $S \approx 0$.

We compute $S$ only when $E_{\mathrm{wrong}} - E_{\mathrm{self}} > 10^{-6}$, that is, when $B$'s own action is better than the wrong actions. $S$ is therefore a conditional score over the pairs that pass this rule. We report how many pairs it removes wherever the errors of single pairs were stored (Section~\ref{sec:eval}). We also report two further measures. Wrong-action accuracy is the share of all pairs in which the correct action beats every wrong action, with a chance level of $1/(1 + |\mathcal{W}|)$. Action sensitivity is the median of $E(B' \rightarrow B) / E(B \rightarrow B)$, where $B'$ is one test clip drawn at random for each target from all test clips, whatever its action class or character. The draw uses a fixed seed, so every method is tested with the same clips. A sensitivity close to 1 suggests that the model makes little use of its action, and in that case we do not interpret its transfer scores.

\subsection{Body-plan gap}

To separate the body from other changes, we take actions from a character of a held-out family and transfer them to three kinds of target. These are another character of the same family, a character of another animal family, and a human. All are drawn in the same style. We compute two differences,
\begin{equation}
\Delta_{\mathrm{animal}} = S_{\mathrm{same\_family}} - S_{\mathrm{other\_family}}, \qquad \Delta_{\mathrm{human}} = S_{\mathrm{same\_family}} - S_{\mathrm{human}} .
\end{equation}
A positive value means that transfer is better within the same family. We give 95\% intervals by resampling target characters, because the frames and pairs of one character are not independent. Each resample draws whole target characters with replacement and keeps all of their pairs. It is not stratified by source family, and source characters are not resampled, so the intervals may be somewhat narrow.

\section{Benchmarks}

We use three sources of video. The synthetic benchmark changes only the body, Sprites changes only the appearance, and the animator-made animations provide motion that was not generated by our own code. Table~\ref{tab:data} gives an overview.

\begin{table}[H]
\centering\small
\caption{The three benchmarks. All frames are used at $64\times64$ pixels.}
\label{tab:data}
\begin{adjustbox}{max width=\textwidth}\begin{tabular}{lccc}
\toprule
 & Synthetic benchmark & Sprites \cite{sprites} & Animator-made animation \cite{dt4d} \\
\midrule
Role & body plan changes, all else fixed & shared body, appearance changes & animator-made 3D motion drawn in 2D \\
Characters & 36 (6 body plans) & 1,296 (1 body plan) & 32 animals in 5 families, 24 humans, 21 other species \\
Actions & run, jump, fall, turn, crouch & walk, spellcard, slash in 3 directions & walk, run, jump, fall, attack, idle \\
Sequences & 3,600 clips & 11,664 clips & 431 animations, each sampled in 3 windows \\
Frames per clip & 16 & 12 & 16 \\
Held out & blob, biped or limbless & 296 test characters & one animal family and all humans \\
\bottomrule
\end{tabular}\end{adjustbox}
\end{table}

\subsection{Synthetic benchmark}

\begin{figure}[!htbp]
\centering
\includegraphics[width=0.8\textwidth]{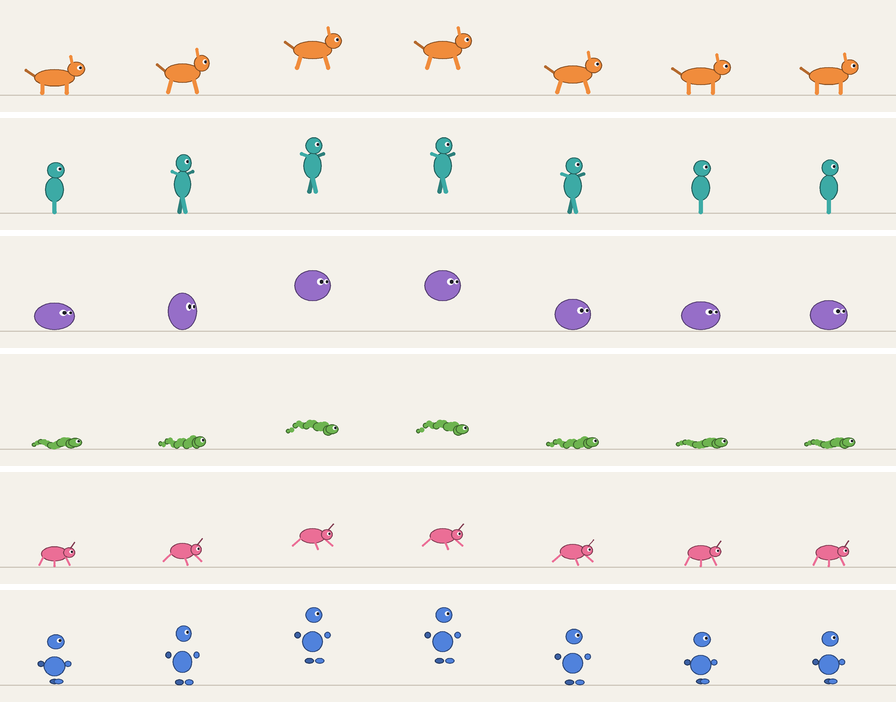}
\caption{One jump on the six body plans, every second frame. From top to bottom, quadruped, biped, blob, limbless, multi-limbed and disjointed.}
\label{fig:jump}
\end{figure}

We propose a new synthetic benchmark for this study, and its availability is described in the Data and code availability section. It has 36 characters. There are six body plans (quadruped, biped, blob, limbless, multi-limbed and disjointed), three shapes for each body plan and two colour schemes. Every character performs five actions (run, jump, fall over, turn and crouch) under 20 intents. An intent fixes the effort, direction, start position and start time of an action. The same intent is used for every action and every character. Because of this, every transfer has a correct answer. If character $A$ jumps under intent $k$, the transferred jump on character $B$ should look like $B$'s own jump under intent $k$. In addition, every clip starts with one to four frames of the character standing still. These frames are the same for all five actions, so the first frame does not give the action away. Fig.~\ref{fig:jump} shows one jump on all six body plans. The quadruped and the biped push off with their legs. The blob squashes and stretches, the limbless body arches, and the parts of the disjointed body fly apart.

Each run holds out one of three body plans, the blob, the biped or the limbless body, and trains on the other five. Training uses intents 0 to 15 of the first two shapes of these five body plans. Testing uses intents 16 to 19, with targets drawn from the third shape, which is never seen, and from the held-out body plan. We train three seeds for each held-out body plan, which gives nine runs per method. The quadruped, multi-limbed and disjointed body plans are always used for training and are never held out.

\subsection{Sprites, a same-body control}

\begin{figure}[!htbp]
\centering
\includegraphics[width=0.78\textwidth]{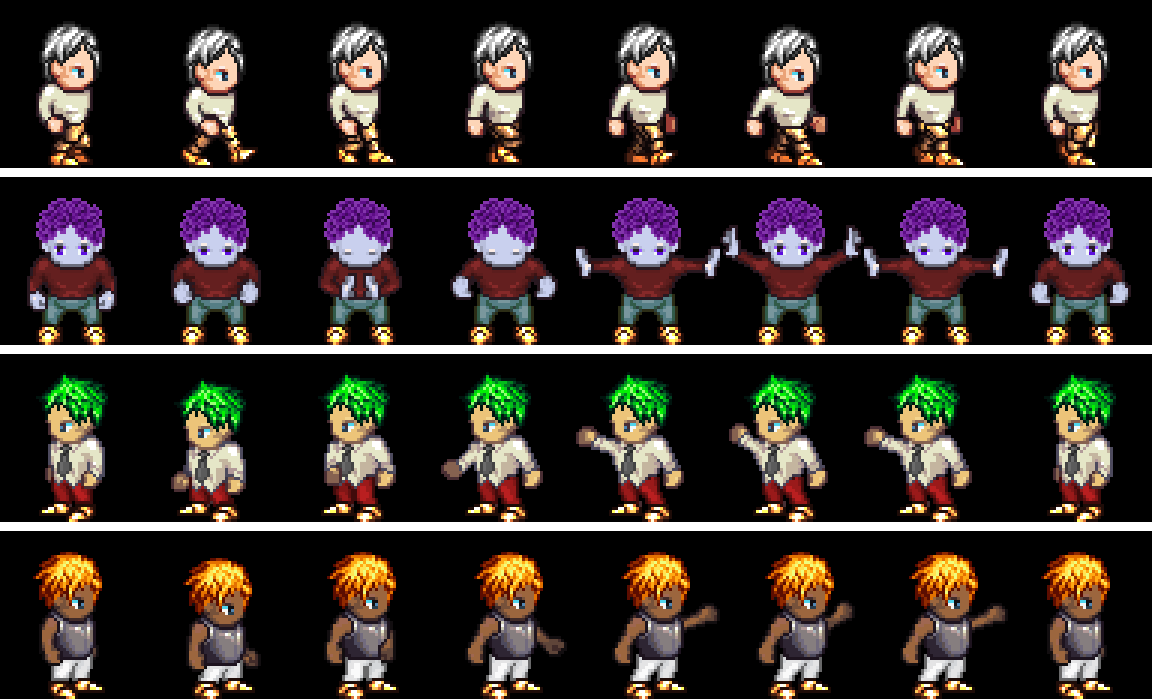}
\caption{Four Sprites characters with the same body and different skin, clothing and hair. From top to bottom, walk to the right, spellcard to the front, slash to the left and slash to the right. Art from the Liberated Pixel Cup, CC BY-SA 3.0.}
\label{fig:sprites}
\end{figure}

Sprites \cite{sprites} has 1,296 characters. They all share one body and differ in skin, top, bottom and hairstyle (Fig.~\ref{fig:sprites}). Each character performs walk, spellcard and slash in three directions. We add one to four standing frames before each action, as in the synthetic benchmark, and use the standard train and test split. Because every character has the same body, Sprites tests transfer across appearance only. It therefore checks that the models can transfer actions at all before the body is changed.

\subsection{Animator-made animation}

\begin{figure}[!htbp]
\centering
\includegraphics[width=0.78\textwidth]{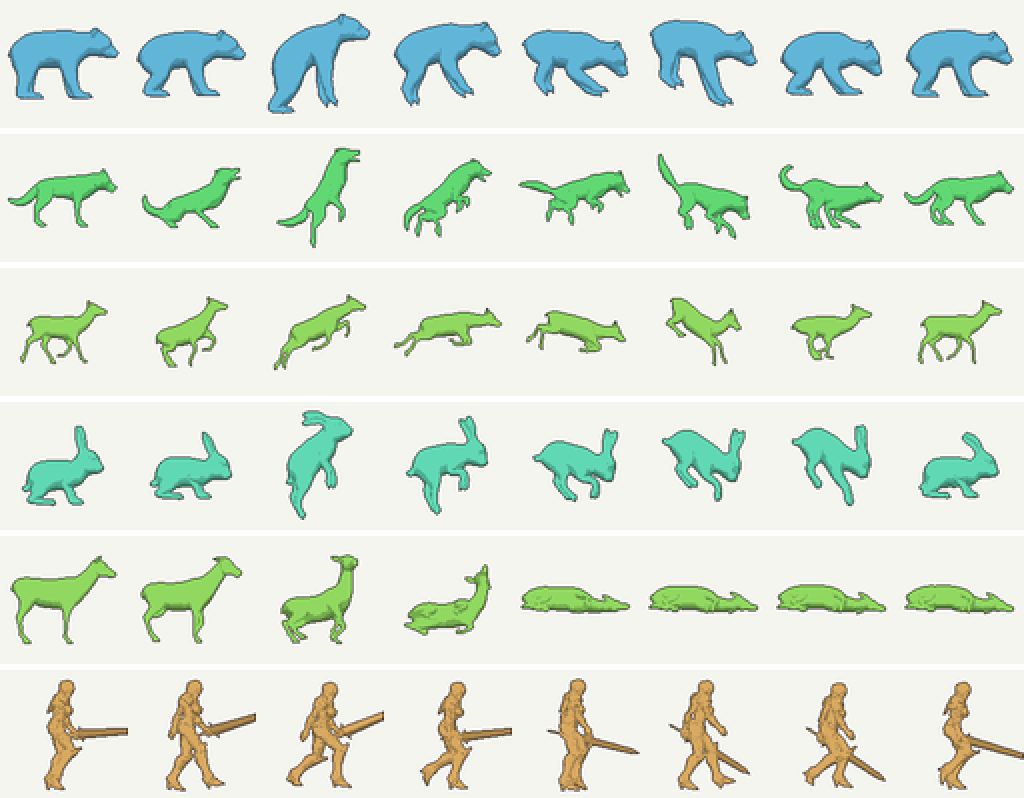}
\caption{Animator-made animations drawn as 2D side views with the body camera, every second frame. From top to bottom, a bear, a dog, a deer and a rabbit jumping, a deer falling and a human walking.}
\label{fig:dt4d}
\end{figure}

DeformingThings4D \cite{dt4d} provides 3D animations of animals and humans. The animal animations were designed by CG artists, and the human animations come from Adobe Mixamo, where they were recorded with motion capture. We render these animations as flat 2D side views so that they can be used with our image-based models. Each character is viewed from the side, drawn in one colour with three shades and a dark outline, and turned to face right (Fig.~\ref{fig:dt4d}). All characters share one visual style, but their body shapes and movements still differ.

We render every animation with two cameras. The fixed camera uses one view for the whole animation. Its frame covers the 0.5th to 99.5th percentile of the character's positions over all frames, with a margin of 12\%, so that nearly all of the motion stays in view. The body camera recentres every frame on the middle of the character's bounding box and keeps one scale for the whole animation. It removes the movement of the character across the image, but changes in size, rotation and shape remain. On the synthetic benchmark, the body camera instead shifts every frame so that the centre of the character's pixels sits in the middle of the image. Neither procedure removes rotation or changes of shape, so we refer to the resulting view as body-centred motion and not as pure body motion.

Our selected subset contains 431 animations. Of these, 169 are labelled animations of 32 animal characters, each with at least four of six action classes. These classes are walk, run, jump, fall, attack and idle. The animals form five families, which are bears, deer, dogs and foxes, rabbits, and a group of cats, raccoons and a hog. A further 235 animations have no usable action label and are used only for training. They include further animations of the 32 labelled animals and animations of 21 other species, such as dragons, birds and a whale. The remaining 27 animations are labelled clips of 24 human characters, which are used only as transfer targets and never in training.

All five labelled animal families walk on four legs. Holding out one family therefore tests transfer to new animal characters with the same broad body plan. Transfer to humans tests a new body plan, although it also brings other differences, such as proportions and clothing. From each animation we sample three clips of 16 evenly spaced frames. One covers the full animation, and the other two cover its first and last three quarters, so they overlap. The two shorter clips are used only for training, and all evaluation uses the full-length clip. Each run holds out one animal family and all humans, and we repeat this for every family. A held-out family is removed from training completely, including its unlabelled animations.

The 21 other species are used only for training and are never held out. However, eight of their animations show felids or a pig (a cat, a leopard, lionesses and a pig), which are related to the cat, raccoon and hog family. This family is held out less strictly than the other four. To test the effect of this overlap, we repeated the main analysis without this family. Every main score, probe and leak value changed by at most 0.03, the order of the methods stayed the same, and the gain of M4 over M2 changed from 0.073 to 0.072. This overlap therefore has little effect on the main comparisons.

\section{Model and Objectives}\label{sec:model}

\subsection{Model}

All methods use the same model (Table~\ref{tab:hp}). The inverse model takes two frames that are $H=3$ steps apart, together with their difference, and maps them to an action vector with four strided convolutions and two linear layers. The action is normalised with batch normalisation without a learned scale. In M2 to M5 we also apply $\tanh$ to the action and add Gaussian noise with a standard deviation of 0.1 during training. The control models in Section~\ref{sec:objectives} switch both on or off together. This limits how much the action can store. The forward model is a U-Net with four levels. The action changes its features through FiLM \cite{film} at the bottleneck and at three decoder levels. The model has 1.85 million parameters with an 8-dimensional action and 1.97 million with a 128-dimensional action.

\subsection{Objectives}\label{sec:objectives}

We compare five model configurations that differ in action size, noise and the losses used to train the action.

M1 uses a 128-dimensional action and only the prediction loss. It is the wide baseline.

M2 uses an 8-dimensional action with noise. M3, M4 and M5 add one loss each to M2.

M3 is a simplified form of CLAW \cite{claw}. In place of a second pass of the world model, a small head tries to rebuild a $16\times16$ copy of the future frame from the action alone. Gradient reversal, which grows to a weight of 0.3 during training, pushes this information out of the action.

M4 is a simplified form of Olaf-World \cite{olaf}. It aligns the action with the feature change between two frames, whereas Olaf-World aligns actions averaged over a whole clip. A projection of the action must point in the same direction as the change in frozen ImageNet ResNet-18 features between the two frames. We use a cosine loss with weight 0.5.

M5 is a simplified form of DiLA \cite{dila}. The action only drives a prediction of the character's silhouette, which is trained with a mask loss. A second network, which never sees the action, paints the appearance onto this silhouette.

M3 to M5 are simplified adaptations of the published methods, not reproductions. Results for M3 to M5 therefore describe these objectives in our setting and not the published systems.

Three control models separate the parts of these objectives. M1 differs from M2 in action size and in noise, so we also train M2 without noise and M1 with noise. M4 adds a loss and also brings in ImageNet features, so we also train M4 with a frozen ResNet-18 that keeps its random initial weights. The two encoders differ only in their weights. Both resize the frames to $112\times112$, use the same ImageNet normalisation and give the same 512 features after global average pooling. The loss and its weight are also the same. In all models, noise and $\tanh$ go together. A model with noise applies $\tanh$ to the action and then adds noise, and a model without noise uses neither. As a result, M2 without noise differs from M1 only in action size, and M1 with noise differs from M2 only in action size. All other settings are unchanged.

\subsection{Displacement baseline}

The baseline has no learned parameters. It moves the target character's pixels by the same amount as the source character's centre moves over the same three frames, and leaves all other pixels unchanged. In this way, it transfers the movement of the source across the screen but not the motion of its body. For this reason, a learned model that does not score higher than this baseline transfers only position.

\section{Training and Evaluation}

\subsection{Training}

All models in the main experiments are trained for 20 epochs with AdamW, a weight decay of $10^{-4}$ and a one-cycle learning rate schedule with a peak of $10^{-3}$. Each batch holds 128 pairs of frames, taken from all start positions in all training clips. The loss is the squared error of the predicted frame, and character pixels count ten times more than background pixels. All models were trained on one GPU on Kaggle, and one model takes a few minutes to train. Table~\ref{tab:hp} lists every setting.

\begin{table}[!htbp]
\centering\small
\caption{Settings shared by all models unless noted.}
\label{tab:hp}
\begin{adjustbox}{max width=\textwidth}\begin{tabular}{ll}
\toprule
Setting & Value \\
\midrule
Frame size & $64\times64$, RGB \\
Prediction horizon $H$ & 3 frames \\
Inverse model & 4 strided convolutions (32, 64, 128, 128 channels), 2 linear layers \\
Forward model & U-Net with 4 levels (32, 64, 128, 128 channels), FiLM from the action \\
Action size & 128 for M1, 8 for M2 to M5 \\
Action noise & standard deviation 0.1 for M2 to M5, none for M1 \\
Parameters & 1.85 million (8-dimensional action), 1.97 million (M1) \\
Optimizer & AdamW, weight decay $10^{-4}$ \\
Learning rate & one-cycle schedule, peak $10^{-3}$ \\
Epochs and batch size & 20 epochs, 128 frame pairs \\
Character pixel weight & 10 \\
M3 adversarial weight & rises to 0.3 \\
M4 alignment weight & 0.5 (cosine loss to ResNet-18 feature change) \\
M5 mask loss & binary cross-entropy on the silhouette \\
\bottomrule
\end{tabular}\end{adjustbox}
\end{table}

\subsection{Evaluation}\label{sec:eval}

For each run we report the median of $S$ over the transfer pairs that pass the rule of Section~\ref{sec:score}, together with $G$, wrong-action accuracy and action sensitivity. Each source animal has two to five other actions, so the chance level of wrong-action accuracy is between 0.17 and 0.33, and the same wrong actions are used for every method. Two linear probes read the action vectors, using their mean and spread over time. The first probe learns the action class from clips of training bodies and is tested on clips of held-out bodies. It measures whether the action class transfers. The second probe predicts the body plan or family of clips from training bodies. It measures how much the action says about its source body, which we call body leak. Both probes are logistic regressions on standardised features with $C=0.5$. On the animator-made animations the action probe uses balanced accuracy, because the classes have different sizes. On the synthetic benchmark it uses plain accuracy. Family leak uses plain accuracy on both benchmarks.

However, probes read the action and not the motion the model makes. To check the generated motion directly, we train a separate convolutional classifier on real clips of the training families. We then drive each held-out animal from one fixed starting pose with the actions of every training animal for five steps, and the classifier labels the result. The starting pose is the same for every action, so only the action can change the label. Its accuracy on real clips of the same animals gives a reference for this test.

$S$ is defined only when $E_{\mathrm{wrong}} - E_{\mathrm{self}}$ is above a small threshold. On the animator-made animations we stored the errors of every pair for every model and checked three thresholds, $10^{-6}$, $10^{-4}$ and $10^{-3}$. No pair of any learned model was removed at any of them. For the baseline, at most 0.7\% of pairs were removed. So the scores of the learned models cover all pairs. The denominator is also far from zero. For every learned model, $E_{\mathrm{wrong}} - E_{\mathrm{self}}$ is at least 20\% of $E_{\mathrm{self}}$ in 95\% of pairs, and its median is between 79\% and 398\%. For the baseline, the 5th percentile is 6\%. The synthetic and Sprites runs used the same threshold of $10^{-6}$, but stored only the median per run, so we cannot give the count there.

Clips of one action class cover the same part of each animation, but two characters may not be at the same point of the action at the same time. For example, the take-off of one jump may be compared with the landing of another. We test the effect of this with a best-shift check. We shift the source action in time by up to three steps in either direction and keep the shift with the lowest error on the target. The shift is chosen with the target's real frames, so this is a diagnostic that gives an upper bound on what timing alignment could add, not a method a model could use. We do this separately for the correct action and for each wrong action, so the correct action gets no advantage. We run this check for M2 and M4 with seed 0 on up to 150 pairs of each kind and family.

The main comparison of all five methods on the animator-made animations uses one seed per held-out family (Sections~\ref{sec:weak} and~\ref{sec:less}). The comparison across the same family, other families and humans (Section~\ref{sec:family}) uses two further seeds of M2 and M4 with all transfer pairs, which gives 10 runs per model. The test of M4 against M2 adds the main seed to these runs, which gives 15 runs. The control models (Section~\ref{sec:less}) use two seeds per family, compared with M2 and M4 trained with the same two seeds, and M1 uses one seed.

\section{Results}

\subsection{The usual gain rewards ignoring the action}

\begin{figure}[!htbp]
\centering
\includegraphics[width=\textwidth]{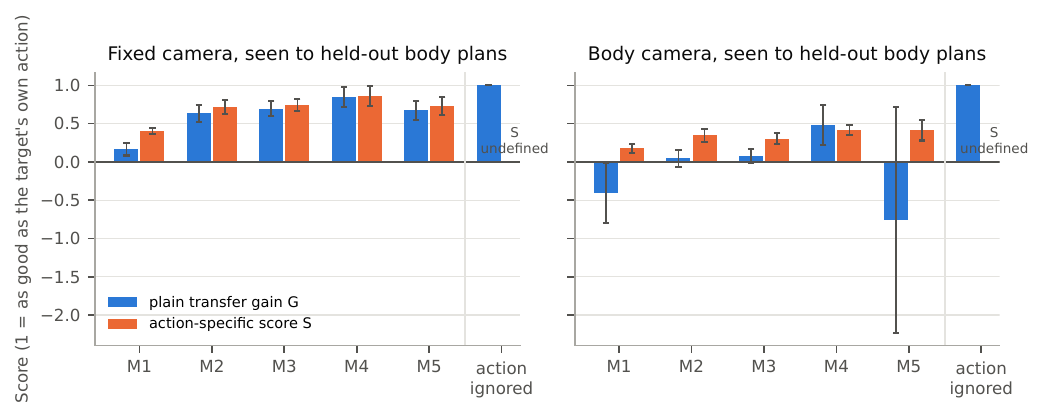}
\caption{The usual gain $G$ and our score $S$ for seen to held-out bodies on the synthetic benchmark, mean and standard deviation over 9 runs. A model that ignores its action (right) gets $G = 1.00$, and its $S$ is undefined.}
\label{fig:metric}
\end{figure}

Fig.~\ref{fig:metric} compares $G$ and $S$ on the synthetic benchmark. A model that ignores its action gets $G=1.00$, higher than every trained model, and its $S$ is undefined. Across the five objectives, $G$ for seen to held-out bodies has a correlation of $r=-0.97$ with action sensitivity under the fixed camera. That is, a model that relies less on its action gets a higher $G$. With the body camera, $G$ is also unstable. M5 gets $G=-0.76 \pm 1.50$ over nine runs, because the body moves little and the difference $E_{\mathrm{copy}}-E_{\mathrm{self}}$ becomes small, and its $S$ is $0.41 \pm 0.13$. We therefore use $S$ for all other results.

\subsection{Transfer works when the body is the same}

On Sprites, every learned model reaches $S$ between 0.98 and 1.00 on new characters (Table~\ref{tab:sprites}). It also finds the correct action class in 100\% of test clips. $S$ stays at 0.97 or higher even when the source and target differ in all four parts of their appearance. This shows that the models learn useful actions, and that a change of appearance alone does not break transfer. The baseline also gets $S = 1.00$, but this value is not informative. Every Sprites character performs an action with the same frames, so the correct source moves its centre almost exactly as the target does, and $E_{\mathrm{tr}}$ is almost equal to $E_{\mathrm{self}}$. The baseline still predicts worse than copying the frame, with a self-gain of $-0.27$.

\begin{table}[!htbp]
\centering\small
\caption{Sprites, one body with new appearances, mean over 3 seeds. Chance is 0.17 for appearance leak.}
\label{tab:sprites}
\begin{adjustbox}{max width=\textwidth}\begin{tabular}{lccccc}
\toprule
Method & Sensitivity & Self-gain & $S$ (new appearance) & Action probe & Appearance leak \\
\midrule
No-learning baseline &  & $-0.27$ & 1.00 &  &  \\
M1 wide action & 27.9 & 0.96 & 0.98 & 1.00 & 1.00 \\
M2 bottleneck & 19.2 & 0.96 & 1.00 & 1.00 & 0.55 \\
M3 CLAW-style & 11.2 & 0.96 & 1.00 & 1.00 & 0.41 \\
M4 Olaf-style & 11.2 & 0.95 & 0.99 & 1.00 & 0.33 \\
M5 DiLA-style & 16.6 & 0.95 & 1.00 & 1.00 & 0.60 \\
\bottomrule
\end{tabular}\end{adjustbox}
\end{table}

\subsection{With a fixed camera, much of the transfer is position}

\begin{figure}[!htbp]
\centering
\includegraphics[width=\textwidth]{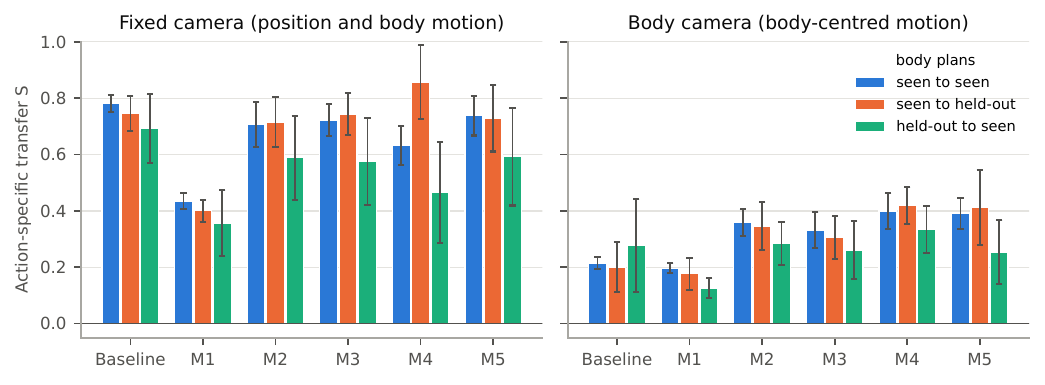}
\caption{Transfer on the synthetic benchmark with the fixed camera (left) and the body camera (right), mean and standard deviation over held-out body plans and seeds.}
\label{fig:tb}
\end{figure}

On the synthetic benchmark with a fixed camera, the baseline gets $S = 0.75 \pm 0.06$ for seen to held-out body plans (Fig.~\ref{fig:tb}, Table~\ref{tab:tb}). In comparison, M2, M3 and M5 score about the same, between 0.72 and 0.74. Only M4 is higher, at 0.86, and M1 is lowest, at 0.40. The animator-made animations show a similar pattern. The baseline gets 0.71 for seen to held-out families, and the learned models get between 0.35 and 0.79 (Fig.~\ref{fig:dt4d_scores}, Table~\ref{tab:dt4d}). A method that only copies movement matches most learned models. Consequently, under a fixed camera, movement across the screen accounts for much of the measured transfer.

\begin{table}[!htbp]
\centering\small
\caption{Synthetic benchmark, mean and standard deviation over 9 runs, or 3 for the baseline. Chance is 0.20 for the probe and for body leak.}
\label{tab:tb}
\begin{adjustbox}{max width=\textwidth}\begin{tabular}{llccccccc}
\toprule
Camera & Method & Sensitivity & $S$ seen$\to$seen & $S$ seen$\to$held-out & $S$ held-out$\to$seen & Accuracy seen$\to$held-out & Probe (held-out) & Body leak \\
\midrule
Fixed & No-learning baseline &  & 0.78 $\pm$ 0.03 & 0.75 $\pm$ 0.06 & 0.69 $\pm$ 0.12 & 0.79 $\pm$ 0.11 &  &  \\
 & M1 wide action & 17.1 & 0.44 $\pm$ 0.03 & 0.40 $\pm$ 0.04 & 0.36 $\pm$ 0.12 & 0.79 $\pm$ 0.08 & 0.81 $\pm$ 0.11 & 0.95 $\pm$ 0.04 \\
 & M2 bottleneck & 10.1 & 0.71 $\pm$ 0.08 & 0.72 $\pm$ 0.09 & 0.59 $\pm$ 0.15 & 0.88 $\pm$ 0.03 & 0.85 $\pm$ 0.13 & 0.80 $\pm$ 0.06 \\
 & M3 CLAW-style & 8.3 & 0.72 $\pm$ 0.06 & 0.74 $\pm$ 0.08 & 0.57 $\pm$ 0.15 & 0.89 $\pm$ 0.04 & 0.78 $\pm$ 0.17 & 0.68 $\pm$ 0.05 \\
 & M4 Olaf-style & 4.9 & 0.63 $\pm$ 0.07 & 0.86 $\pm$ 0.13 & 0.47 $\pm$ 0.18 & 0.80 $\pm$ 0.05 & 0.55 $\pm$ 0.20 & 0.77 $\pm$ 0.04 \\
 & M5 DiLA-style & 10.7 & 0.74 $\pm$ 0.07 & 0.73 $\pm$ 0.12 & 0.59 $\pm$ 0.17 & 0.85 $\pm$ 0.03 & 0.86 $\pm$ 0.09 & 0.75 $\pm$ 0.05 \\
\midrule
Body & No-learning baseline &  & 0.22 $\pm$ 0.02 & 0.20 $\pm$ 0.09 & 0.28 $\pm$ 0.16 & 0.30 $\pm$ 0.02 &  &  \\
 & M1 wide action & 14.0 & 0.20 $\pm$ 0.02 & 0.18 $\pm$ 0.06 & 0.13 $\pm$ 0.03 & 0.42 $\pm$ 0.06 & 0.54 $\pm$ 0.18 & 0.99 $\pm$ 0.00 \\
 & M2 bottleneck & 7.6 & 0.36 $\pm$ 0.05 & 0.35 $\pm$ 0.08 & 0.28 $\pm$ 0.08 & 0.52 $\pm$ 0.04 & 0.62 $\pm$ 0.11 & 0.85 $\pm$ 0.04 \\
 & M3 CLAW-style & 7.1 & 0.33 $\pm$ 0.06 & 0.30 $\pm$ 0.08 & 0.26 $\pm$ 0.10 & 0.49 $\pm$ 0.03 & 0.60 $\pm$ 0.17 & 0.73 $\pm$ 0.05 \\
 & M4 Olaf-style & 2.9 & 0.40 $\pm$ 0.06 & 0.42 $\pm$ 0.07 & 0.33 $\pm$ 0.08 & 0.47 $\pm$ 0.06 & 0.48 $\pm$ 0.10 & 0.81 $\pm$ 0.03 \\
 & M5 DiLA-style & 8.2 & 0.39 $\pm$ 0.06 & 0.41 $\pm$ 0.13 & 0.25 $\pm$ 0.11 & 0.49 $\pm$ 0.04 & 0.57 $\pm$ 0.15 & 0.79 $\pm$ 0.05 \\
\bottomrule
\end{tabular}\end{adjustbox}
\end{table}

\subsection{Body motion transfers weakly}\label{sec:weak}

\begin{figure}[!htbp]
\centering
\includegraphics[width=\textwidth]{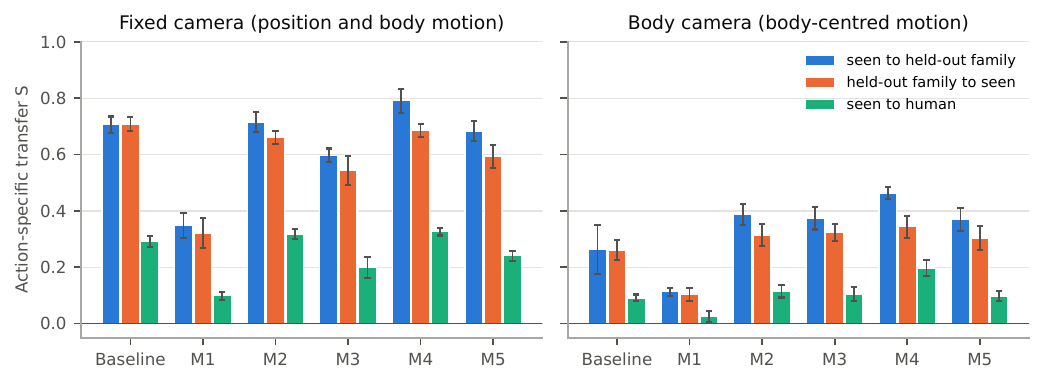}
\caption{Transfer on the animator-made animations, mean and standard deviation over the 5 held-out families.}
\label{fig:dt4d_scores}
\end{figure}

With the body camera, the baseline has no movement to copy, and its score on the synthetic benchmark falls to 0.20. The learned models also fall, to between 0.18 (M1) and 0.42 (M4) for seen to held-out body plans. On the animator-made animations, the best model reaches 0.46 for a held-out animal family, and every model scores at most 0.20 from seen bodies to humans (Fig.~\ref{fig:dt4d_scores}). Without the less strictly held-out cat, raccoon and hog family, these values are 0.47 and 0.20. For sources from the seen families, changing the target from a held-out animal family to a human lowers the score of M2 to M5 by 0.26 to 0.27. Because some human clips were drawn from the front in a first run, we redrew all humans from the side, facing right like the animals, and trained and tested every model again. The human scores with the body camera changed by at most 0.01, so the drop to humans does not come from the viewpoint.

The result also depends on the action. With the body camera on the synthetic benchmark, the baseline gets 0.81 for turn and 0.54 for crouch, which mostly change the outline. In contrast, it gets $-0.19$ for run and $-0.22$ for jump, which need the legs or the whole body to move in a way that depends on the body. The learned models, on the other hand, get 0.10 to 0.25 on run and 0.15 to 0.34 on jump.

\begin{table}[!htbp]
\centering\small
\caption{Animator-made animations, mean and standard deviation over the 5 held-out families. The probe is balanced accuracy and family leak is plain accuracy. Chance is 0.17 for the probe and 0.25 for family leak.}
\label{tab:dt4d}
\begin{adjustbox}{max width=\textwidth}\begin{tabular}{llccccccc}
\toprule
Camera & Method & Sensitivity & $S$ seen$\to$held-out & $S$ held-out$\to$seen & $S$ seen$\to$human & Accuracy seen$\to$held-out & Probe (held-out family) & Family leak \\
\midrule
Fixed & No-learning baseline &  & 0.71 $\pm$ 0.03 & 0.71 $\pm$ 0.02 & 0.29 $\pm$ 0.02 & 0.75 $\pm$ 0.04 &  &  \\
 & M1 wide action & 3.9 & 0.35 $\pm$ 0.05 & 0.32 $\pm$ 0.05 & 0.10 $\pm$ 0.01 & 0.65 $\pm$ 0.04 & 0.88 $\pm$ 0.08 & 0.74 $\pm$ 0.04 \\
 & M2 bottleneck & 2.1 & 0.71 $\pm$ 0.03 & 0.66 $\pm$ 0.02 & 0.32 $\pm$ 0.02 & 0.69 $\pm$ 0.04 & 0.87 $\pm$ 0.07 & 0.45 $\pm$ 0.05 \\
 & M3 CLAW-style & 1.9 & 0.60 $\pm$ 0.02 & 0.54 $\pm$ 0.05 & 0.20 $\pm$ 0.04 & 0.60 $\pm$ 0.04 & 0.74 $\pm$ 0.13 & 0.39 $\pm$ 0.04 \\
 & M4 Olaf-style & 1.7 & 0.79 $\pm$ 0.04 & 0.68 $\pm$ 0.02 & 0.33 $\pm$ 0.01 & 0.70 $\pm$ 0.04 & 0.83 $\pm$ 0.07 & 0.34 $\pm$ 0.03 \\
 & M5 DiLA-style & 2.1 & 0.68 $\pm$ 0.04 & 0.59 $\pm$ 0.04 & 0.24 $\pm$ 0.02 & 0.70 $\pm$ 0.05 & 0.87 $\pm$ 0.08 & 0.48 $\pm$ 0.05 \\
\midrule
Body & No-learning baseline &  & 0.26 $\pm$ 0.09 & 0.26 $\pm$ 0.04 & 0.09 $\pm$ 0.01 & 0.43 $\pm$ 0.03 &  &  \\
 & M1 wide action & 3.9 & 0.11 $\pm$ 0.01 & 0.10 $\pm$ 0.02 & 0.02 $\pm$ 0.02 & 0.47 $\pm$ 0.06 & 0.81 $\pm$ 0.07 & 0.95 $\pm$ 0.01 \\
 & M2 bottleneck & 1.9 & 0.39 $\pm$ 0.04 & 0.31 $\pm$ 0.04 & 0.12 $\pm$ 0.02 & 0.71 $\pm$ 0.01 & 0.81 $\pm$ 0.07 & 0.47 $\pm$ 0.07 \\
 & M3 CLAW-style & 1.8 & 0.37 $\pm$ 0.04 & 0.32 $\pm$ 0.03 & 0.11 $\pm$ 0.02 & 0.70 $\pm$ 0.02 & 0.72 $\pm$ 0.14 & 0.38 $\pm$ 0.07 \\
 & M4 Olaf-style & 1.5 & 0.46 $\pm$ 0.02 & 0.34 $\pm$ 0.04 & 0.20 $\pm$ 0.03 & 0.69 $\pm$ 0.04 & 0.66 $\pm$ 0.13 & 0.38 $\pm$ 0.06 \\
 & M5 DiLA-style & 1.9 & 0.37 $\pm$ 0.04 & 0.30 $\pm$ 0.04 & 0.10 $\pm$ 0.02 & 0.70 $\pm$ 0.04 & 0.86 $\pm$ 0.09 & 0.53 $\pm$ 0.05 \\
\bottomrule
\end{tabular}\end{adjustbox}
\end{table}

\subsection{Same family, other families and humans}\label{sec:family}

\begin{figure}[!htbp]
\centering
\includegraphics[width=0.6\textwidth]{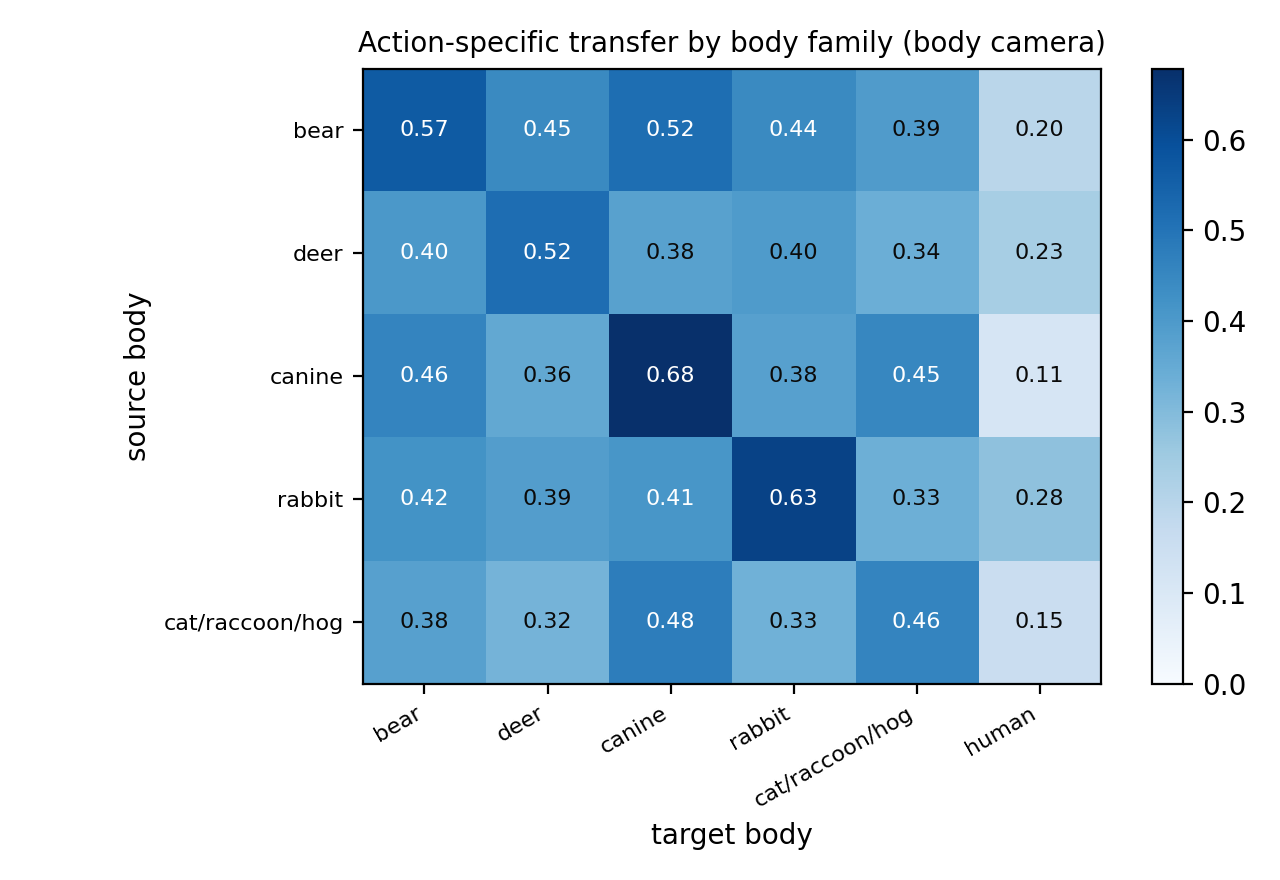}
\caption{Body-centred transfer of M4 from each held-out source family (rows) to each target family (columns). The diagonal is transfer within a family.}
\label{fig:matrix}
\end{figure}

To separate the body from the drawing style, we compare transfer from a character of the held-out family to three kinds of target, drawn in the same style. The target is another character of the same family, a character of another animal family, or a human (Table~\ref{tab:gap}, Fig.~\ref{fig:matrix}). Transfer decreases as the target becomes less similar to the source. For M4, it falls from 0.55 within a family to 0.35 for another animal family and 0.20 for humans. Similarly, for M2 it falls from 0.49 to 0.33 and 0.12. The advantage of the same family over other animal families, and over humans, is positive in all 10 runs of both models. The baseline also does better within a family (0.45) than across families (0.27), so part of the benefit of a similar body comes from a similar outline. The learned models add 0.04 to 0.12 over the baseline in every case. The gap from the same family to humans, however, is the same for all three methods (0.38 to 0.39). M4 raises every score but does not shrink the gap to a different body plan.

A simple offset in time does not explain the gap (Table~\ref{tab:phase}). Even when the best shift is chosen with the target's real frames, transfer from the held-out family to humans rises from 0.10 to 0.17 for M2 and from 0.20 to 0.28 for M4. Humans gain more from the shift than animals do, so differences in timing explain part of the low human scores. The gap from the same family to humans changes only from 0.42 to 0.39 for M2 and from 0.39 to 0.35 for M4.

\begin{table}[!htbp]
\centering\small
\caption{Body-centred transfer $S$ from the held-out family to each kind of target, median with a 95\% interval over target characters.}
\label{tab:gap}
\begin{adjustbox}{max width=\textwidth}\begin{tabular}{lcccc}
\toprule
Method & \makecell{Same\\family} & \makecell{Other quadruped\\family} & Human & \makecell{Gap to\\human} \\
\midrule
No-learning baseline & 0.45 [0.39, 0.54] & 0.27 [0.23, 0.31] & 0.08 [0.01, 0.15] & 0.39 [0.32, 0.47] \\
M2 bottleneck & 0.49 [0.43, 0.55] & 0.33 [0.30, 0.35] & 0.12 [0.04, 0.20] & 0.38 [0.32, 0.45] \\
M4 Olaf-style & 0.55 [0.50, 0.61] & 0.35 [0.33, 0.37] & 0.20 [0.10, 0.31] & 0.38 [0.32, 0.45] \\
\bottomrule
\end{tabular}
\end{adjustbox}
\end{table}

\begin{table}[!htbp]
\centering\small
\caption{Standard scoring against the best time shift of up to three steps, seed 0 and a subsample of pairs, mean over the five families.}
\label{tab:phase}
\begin{adjustbox}{max width=\textwidth}\begin{tabular}{llccccc}
\toprule
Model & Scoring & \makecell{Held-out\\$\to$same family} & \makecell{Held-out\\$\to$other family} & \makecell{Held-out\\$\to$human} & \makecell{Seen\\$\to$human} & \makecell{Gap to\\human} \\
\midrule
M2 & standard & 0.52 & 0.33 & 0.10 & 0.11 & 0.42 \\
M2 & best shift & 0.56 & 0.36 & 0.17 & 0.18 & 0.39 \\
M4 & standard & 0.59 & 0.36 & 0.20 & 0.21 & 0.39 \\
M4 & best shift & 0.63 & 0.37 & 0.28 & 0.31 & 0.35 \\
\bottomrule
\end{tabular}
\end{adjustbox}
\end{table}

\subsection{What transfers, but not how}

\begin{figure}[!htbp]
\centering
\includegraphics[width=\textwidth]{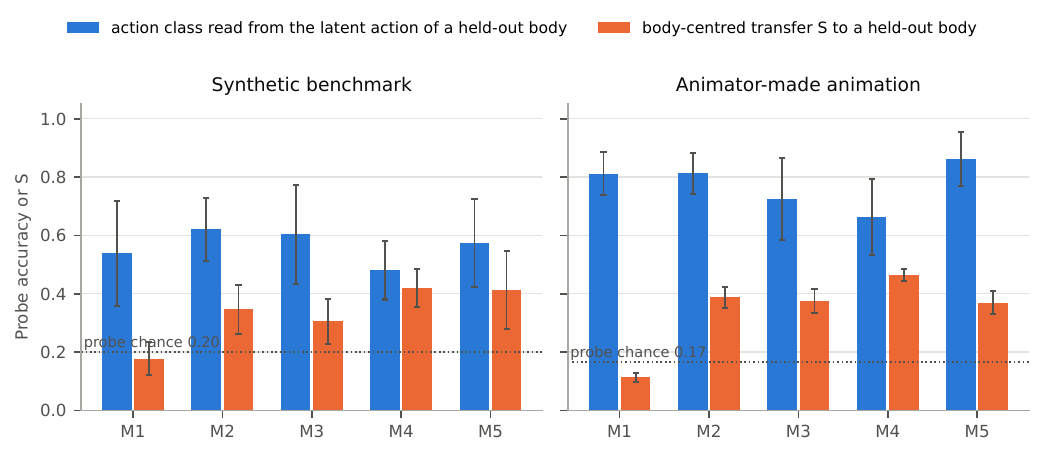}
\caption{Action class read by a linear probe (blue) and body-centred transfer $S$ (orange) for held-out bodies. Probe values are balanced accuracy on the animator-made animations and plain accuracy on the synthetic benchmark. The dotted line is the probe's chance level.}
\label{fig:whathow}
\end{figure}

A linear classifier still identifies the action class from the actions of unseen bodies. With the body camera on the animator-made animations, a linear probe trained on the actions of the training families identifies the action of a held-out family with 66\% to 86\% balanced accuracy, where chance is 17\%. However, for the same models, body-centred transfer is only 0.11 to 0.46 (Fig.~\ref{fig:whathow}). With the body camera on the synthetic benchmark, the probe reaches 48\% to 62\% plain accuracy, where chance is 20\%, and transfer is 0.18 to 0.42. The action class can still be identified from the actions of unseen bodies, but this does not mean that the model can reproduce how those bodies perform the action.

We also use the separate classifier to check the generated motion. When a held-out animal is driven by another animal's action, the classifier finds the intended action in 52\% (M2) and 45\% (M4) of cases. These rates are close to those when the animal is driven by its own action (49\% and 41\%). On real clips of the same held-out families, it recognises 81\%, where chance is 17\%. Because the classifier was trained only on animals, it recognises only 31\% of real human clips, so we do not use it for humans.

\begin{figure}[!htbp]
\centering
\includegraphics[width=0.56\textwidth]{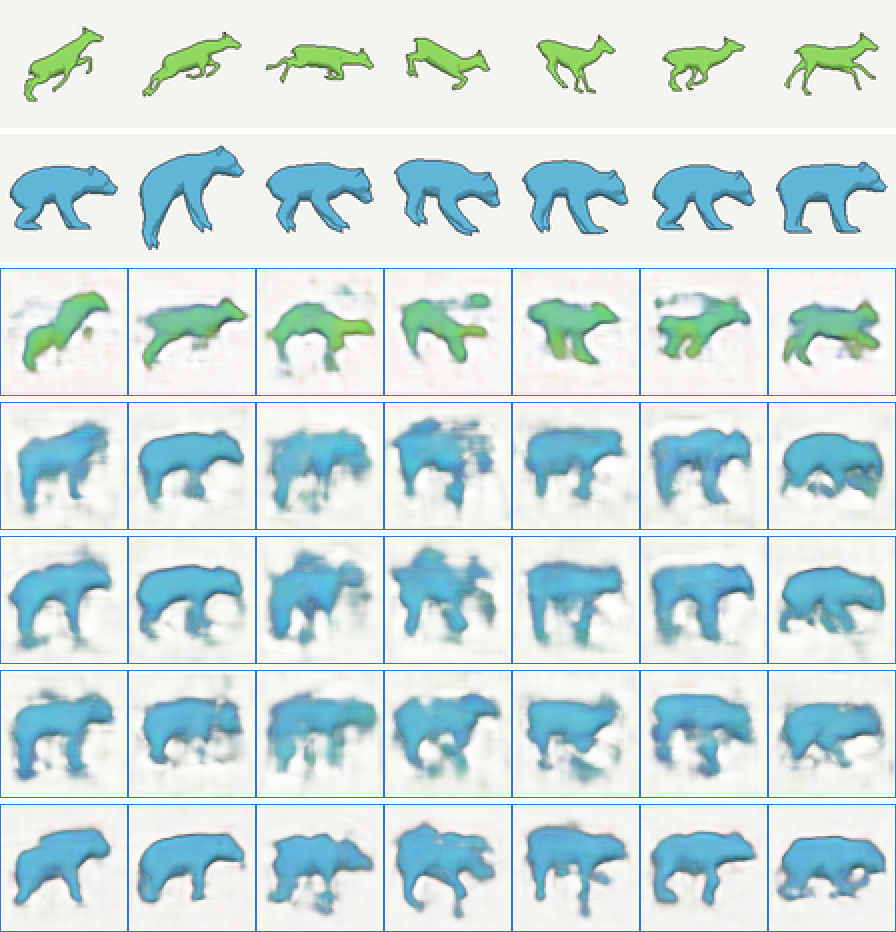}\\[4pt]
\includegraphics[width=0.56\textwidth]{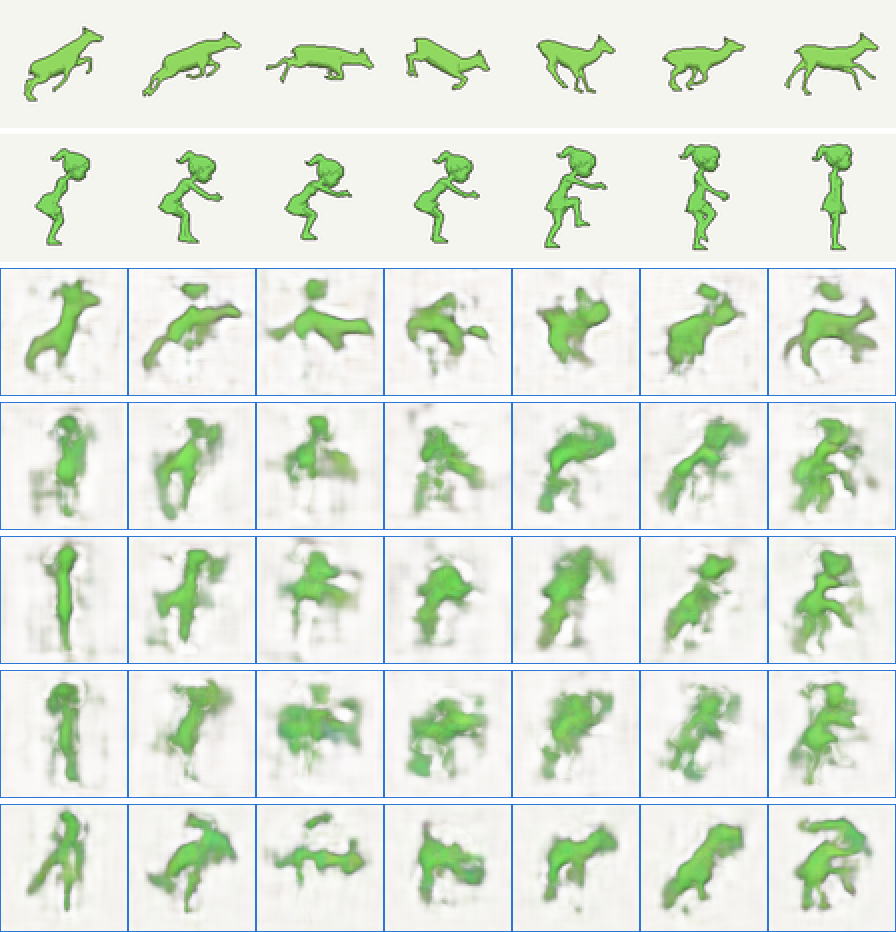}
\caption{A deer's jump transferred to a bear (top) and a human (bottom). From top to bottom, the source, the target's real jump, and the predictions of M1 to M5.}
\label{fig:qual}
\end{figure}

Fig.~\ref{fig:qual} shows the same gap in examples. When the models get a deer's jump, they keep the bear's shape and colour, but they predict a blurred bear instead of a bear leaving the ground. With the same jump, the human target is partly redrawn as a four-legged shape, most clearly by M1 and M5. The transferred action carries part of the shape of the source body.

\subsection{Less body in the action, better transfer}\label{sec:less}

\begin{figure}[!htbp]
\centering
\includegraphics[width=\textwidth]{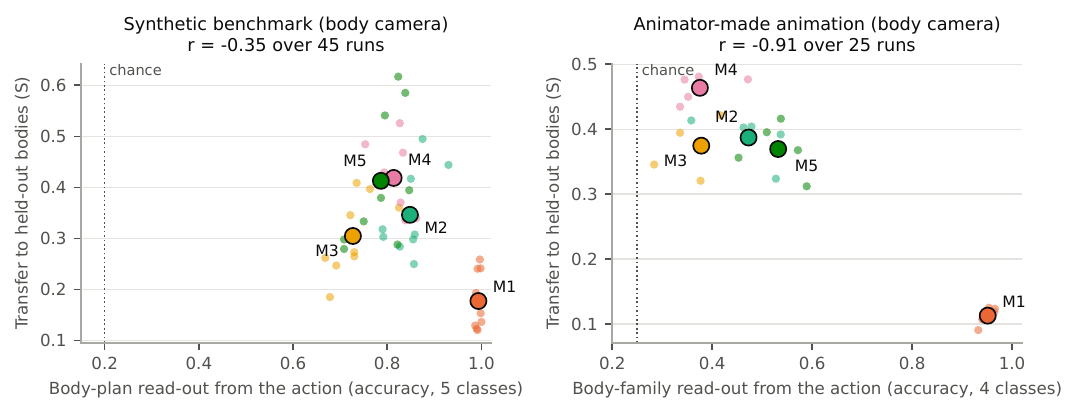}
\caption{Body leak in the action (x axis) against body-centred transfer (y axis). Small points are runs and large points are model means.}
\label{fig:leak}
\end{figure}

The largest difference between the models is the amount of body information in their actions. With the body camera, a classifier identifies the training body plan from the 128-dimensional action of M1 in 99\% of synthetic clips and the body family in 95\% of animator-made clips, and M1 transfers worst of all learned models on every benchmark. The 8-dimensional action of M2, which also adds noise, lowers family leak to 47\% and raises body-centred transfer over M1 by 0.27 on average (paired over the five held-out families), and in every family. The control models show that the size of the action causes most of this (Table~\ref{tab:controls}). Without noise on either model, the 8-dimensional action scores 0.23 higher than the 128-dimensional one, in all 5 families. With noise on both, it scores 0.24 higher, in all 10 runs. Over the paired runs, noise adds on average only 0.05 at 8 dimensions and 0.03 at 128 dimensions. In the action-size controls, family leak follows the size as well, 94\% to 95\% with 128 dimensions and 49\% to 57\% with 8. Among the five main objectives, M3 and M4 have the lowest family leak, both 38\%, and M4 has the highest transfer. However, on the synthetic benchmark M3 has lower body leak than M4 (73\% against 81\%), so low leak alone does not explain the transfer of M4. Over the 25 animator-made runs, family leak and body-centred transfer have a correlation of $r=-0.91$ (Fig.~\ref{fig:leak}). These runs share methods, folds and training data, so they are not independent, and the correlation does not establish a cause. On the synthetic benchmark, though, the correlation is weaker, $r=-0.35$ over 45 runs, because all models except M3 still tell the body plan in 79\% to 99\% of clips.

On the animator-made animations, M4 gives the highest body-centred transfer scores $S$ of the five objectives. For seen to held-out bodies with the body camera on the animator-made animations, M4 scores higher than M2 in all 15 paired runs (3 seeds and 5 families). The mean gain is 0.073, and single runs range from 0.028 to 0.120. Averaged per family, the gain is 0.052 for bears, 0.064 for dogs and foxes, 0.079 for rabbits, 0.079 for cats, raccoons and the hog, and 0.091 for deer. A $t$ interval over the five family means gives $[0.054, 0.092]$. Only the five families are independent groups, and they share part of their training data, so this shows a consistent gain in our experiments and not a general advantage of M4. A paired test over the five family means gives $t(4)=10.6$. On the synthetic benchmark, M4 scores 0.072 higher in 8 of 9 runs ($t(8)=2.82$, $p=0.023$, $d_z=0.94$). In contrast, for seen to held-out transfer on the animator-made animations, neither M3 nor M5 scores higher than M2. With the body camera, M4 also has the lowest action sensitivity, 1.5 against 1.9 for M2 (Table~\ref{tab:dt4d}), and the lowest class probe accuracy of the 8-dimensional models. So its gain does not come from a more detailed action.

To find out whether the gain comes from the extra loss or from the ImageNet features, we replace the frozen encoder with a ResNet-18 that has random weights (Table~\ref{tab:controls}). The result differs between animal and human targets. For seen to held-out animal families, the random encoder adds only 0.01 over M2 (bootstrap interval over the 10 runs, $[-0.004, 0.025]$). Across the 10 paired runs, the ImageNet encoder scores on average 0.066 higher than the random encoder, and it is higher in every run. For humans, the random encoder does as well as the ImageNet encoder, 0.20 for both, and both are 0.08 above M2 in all 10 runs. The random encoder also gives the lowest family leak, 36\%, but transfers worse to animals than M4. For new animal families, M4 with the ImageNet encoder scores 0.075 above M2, and M4 with the random encoder only 0.01. This suggests that the pretrained features help, although we did not test why. For humans, the gain comes from the extra loss itself. A random convolutional network still gives features that change when the image changes, mostly with coarse shape and colour. The loss encourages a projection of the action to align with this change, so the action may capture the overall change of the frame. This may favour coarse changes that many bodies share over details of the source body.

\begin{table}[!htbp]
\centering\small
\caption{Control models, mean and standard deviation over runs. They separate action size from noise, and the extra loss of M4 from its ImageNet features.}
\label{tab:controls}
\begin{adjustbox}{max width=\textwidth}\begin{tabular}{lcccccccc}
\toprule
Model & \makecell{Action\\size} & Noise & \makecell{Frozen\\encoder} & Runs & \makecell{$S$ seen\\$\to$held-out} & \makecell{$S$ held-out\\$\to$other family} & \makecell{$S$ seen\\$\to$human} & \makecell{Family\\leak} \\
\midrule
M1 & 128 & no & none & 5 & 0.11 $\pm$ 0.01 & 0.10 $\pm$ 0.02 & 0.03 $\pm$ 0.01 & 0.95 $\pm$ 0.01 \\
M1 with noise & 128 & yes & none & 10 & 0.15 $\pm$ 0.01 & 0.13 $\pm$ 0.03 & 0.05 $\pm$ 0.00 & 0.94 $\pm$ 0.01 \\
M2 without noise & 8 & no & none & 10 & 0.34 $\pm$ 0.03 & 0.28 $\pm$ 0.04 & 0.11 $\pm$ 0.01 & 0.57 $\pm$ 0.08 \\
M2 & 8 & yes & none & 10 & 0.39 $\pm$ 0.04 & 0.33 $\pm$ 0.04 & 0.13 $\pm$ 0.02 & 0.49 $\pm$ 0.06 \\
M4 random encoder & 8 & yes & random & 10 & 0.40 $\pm$ 0.03 & 0.29 $\pm$ 0.02 & 0.20 $\pm$ 0.03 & 0.36 $\pm$ 0.05 \\
M4 & 8 & yes & ImageNet & 10 & 0.47 $\pm$ 0.03 & 0.35 $\pm$ 0.03 & 0.20 $\pm$ 0.02 & 0.38 $\pm$ 0.04 \\
\bottomrule
\end{tabular}
\end{adjustbox}
\end{table}

\section{Discussion}

\subsection{Three parts of an action}

The usual score combines three parts of an action. Our tests separate them for the models and data we use. A baseline that copies only the movement across the screen scores as high as most learned models, so a high score with a fixed camera does not show that an action was learned. The action class can still be read from the actions of unseen bodies. However, the body-centred score of M4 drops from 0.55 within a family to 0.20 for humans. A latent action learned from pixels can therefore keep the action class without giving the forward model enough information to reproduce the motion of the target body. Garrido et al.~\cite{garrido2026} also found that actions learned without a shared body can become tied to the camera. Our baseline is a related test, because it measures how much transfer can be achieved through movement on the screen alone.

\subsection{Should an action be free of the body?}

The link between body leak and transfer suggests that reducing body information in the action can improve transfer. The control models support this for action size, since a smaller action both leaks less and transfers better. However, low body leak does not explain the result by itself, because M4 with a random encoder leaks the least and still transfers worse to animals than M4. The gain of M4 also comes from different sources for the two kinds of target. For new animal families the ImageNet features matter, which suggests that pretrained knowledge of how animals look helps when the target is also an animal. For human targets the random encoder does as well, so pretrained features add nothing there, and the gain comes from the extra loss. Still, the gap to humans is the same for M2, M4 and the baseline. The best model closes 46\% of the error gap between wrong actions and the target's own action for a new animal family, and at most 20\% for humans. An action with less information about the source body may still indicate that a jump is happening. The forward model must then work out how the new body jumps, although it has only seen the training bodies jump. We have not tested this idea, but a future model could describe the target body separately from the action and write motion on a new body as
\begin{equation}
x_{t+H} = f(x_t,\, a,\, \mathrm{body}),
\end{equation}
with the action $a$ and the body as separate inputs. In current latent-action models, by contrast, the body is only hidden inside $x_t$. One option is a model that describes the structure of the target directly, for example as parts learned from its own video \cite{apes}, and applies the action to these parts. In such a model, the action would say which parts move and in which direction, for example that the support parts push down so that the whole body rises. The body description would then say which parts of the new character play these roles. Transfer would then require matching parts between bodies. This is hard for a blob without legs, and our benchmark can test it.

\subsection{What this means for evaluation}

We recommend that work on transfer across bodies includes a baseline that copies only the movement on the screen and a score that compares the correct action with wrong ones. The usual gain can rank models in the reverse order of how much they use the action. Transfer between bodies with a similar structure, such as two robot arms, is not evidence of transfer between different body plans, because the body-specific motion is the part that transfers least in our tests.

\section{Limitations}
\label{sec:lim}

We use small models at $64\times64$ pixels, all with the same training settings. Between M2 and M3 to M5 only the objective changes. M1 also differs in action size and noise, and we use the control models to separate these two factors. Larger models trained on more varied video may transfer better, so this needs to be tested before our results are applied to large world models. M3 to M5 are adaptations of published methods to our model and may behave differently from the original systems. We also do not compare with skeleton-based motion retargeting \cite{aberman2020,anytop}, because it needs a rig for every character, and most 2D footage does not have one. Our best-shift check only shifts whole action sequences in time. It does not handle actions that are faster or slower on one body than on another. There are only 32 labelled animal characters and 24 human characters. In addition, $S$ measures the inverse and forward models together, so a low score can come from a forward model that cannot draw the new body as well as from the action. The body-centred view still uses pixel error and does not measure joints, contact or body parts directly. The synthetic and Sprites runs stored only the median score per run, so we cannot report how many pairs the condition on $S$ removed there. The synthetic characters move by hand-set rules. The animator-made animations fix this, but they come from 3D meshes drawn in a flat style, without the squash, stretch and smear of hand-drawn cartoons. On the animator-made animations, M2 and M4 use three seeds per held-out family, the control models use two, and M1, M3 and M5 use one. The 24 human characters differ from the animals in more than the body plan, for example in proportions and clothing. Their motion was also recorded with motion capture, whereas the animal motion was made by artists. The human test therefore measures a change of body plan together with these differences. A small test on 43 clips from public-domain cartoons of 1914 to 1916, all checked by the author, gave no clear result. No method distinguished walking, running and falling better than chance, including a frozen ImageNet encoder and a motion baseline. We attribute this to the poor image quality and the small number of clips.

\section{Conclusion}

In the settings we test, a linear classifier still identifies the action class from the latent actions of unseen animal families and held-out synthetic body plans, but the transferred action closes only part of the prediction error gap on these bodies. On synthetic characters with six body plans and on animator-made animations of five animal families and humans, a baseline that only copies movement matches most learned models under a fixed camera. Body-centred transfer reaches at most 0.46 for a new animal family and at most 0.20 for humans, or 0.31 from seen bodies to humans when the best time shift is chosen with the target's frames (Table~\ref{tab:phase}). In our experiments, transfer is better when the action has fewer dimensions and holds less information about the source body. Neither a better objective nor allowing for differences in timing closes the gap to human targets, which differ from the animals in body plan, proportions, clothing and the way their motion was made. For this reason, transfer should be measured with a movement baseline and with a score that compares the correct action against wrong ones, rather than with the usual gain alone. Good transfer between similar bodies also did not carry over to bodies with a different structure. Future models could describe the target body separately from the transferred action, and testing whether this closes the gap is a natural next step.

\section*{CRediT authorship contribution statement}
\textbf{Amarnath R}. Conceptualization, Methodology, Software, Validation, Formal analysis, Investigation, Data curation, Visualization, Writing (original draft), Writing (review and editing).

\section*{Declaration of competing interest}
The author declares that he has no known competing financial interests or personal relationships that could have appeared to influence the work reported in this paper.

\section*{Funding}
This research did not receive any specific grant from funding agencies in the public, commercial, or not-for-profit sectors.

\section*{Data and code availability}
The synthetic benchmark proposed in this study is available on Hugging Face under the CC BY 4.0 licence at
\begin{center}\url{https://huggingface.co/datasets/amarnathresearch/synthetic-2d-body-plans}\end{center} The code that generates the benchmark, the code that renders the animator-made animations, and all training and evaluation code will be released on publication. DeformingThings4D is available from its authors under their terms of use, and Sprites under CC BY-SA 3.0 and GPL 3.0.

\raggedright

\end{document}